\pdfoutput=1

\documentclass[11pt]{article}

\usepackage[]{acl}

\usepackage{times}
\usepackage{latexsym}

\usepackage[T1]{fontenc}

\usepackage[utf8]{inputenc}

\usepackage{microtype}

\usepackage{inconsolata}
\usepackage{xspace}
\newcommand{\method}{\texttt{PILOT}\xspace}
\usepackage{enumitem}
\usepackage{booktabs}
\usepackage{multirow}
\usepackage{stfloats}
\usepackage{amsmath}
\usepackage{amssymb}
\usepackage{mathtools}
\usepackage{amsthm}
\newtheorem{problem}{Problem}

\title{\method: Legal Case Outcome Prediction with Case Law}

\author{Lang Cao\textsuperscript{\rm 1}, 
Zifeng Wang\textsuperscript{\rm 1}, Cao Xiao\textsuperscript{\rm 2}, Jimeng Sun\textsuperscript{\rm 1} \\
\textsuperscript{\rm 1}University of Illinois Urbana-Champaign \\
\textsuperscript{\rm 2}GE Healthcare \\
\texttt{\{langcao2, zifengw2, jimeng\}@illinois.edu, cao.xiao@ge.com}}

\begin{document}
\maketitle
\begin{abstract}
Machine learning shows promise in predicting the outcome of legal cases, but most research has concentrated on civil law cases rather than case law systems. We identified two unique challenges in making legal case outcome predictions with case law. First, it is crucial to identify relevant precedent cases that serve as fundamental evidence for judges during decision-making. Second, it is necessary to consider the evolution of legal principles over time, as early cases may adhere to different legal contexts. In this paper, we proposed a new framework named PILOT (PredictIng Legal case OuTcome) for case outcome prediction. It comprises two modules for relevant case retrieval and temporal pattern handling, respectively. To benchmark the performance of existing legal case outcome prediction models, we curated a dataset from a large-scale case law database. We demonstrate the importance of accurately identifying precedent cases and mitigating the temporal shift when making predictions for case law, as our method shows a significant improvement over the prior methods that focus on civil law case outcome predictions.
\end{abstract}

\section{Introduction}

Predicting legal case outcomes is a crucial task that facilitates data-driven decision-making in legal cases based on relevant information, such as the factual description \cite{cui_survey_2022}. With a significant number of legal cases arising worldwide each year, legal professionals face the daunting task of reviewing the extensive legal text and delivering accurate and fair outcomes. Legal case outcome prediction has the potential to simplify this labor-intensive document review process, enhancing strategy and decision-making. As the volume and complexity of cases continue to escalate, the development of precise and reliable legal case outcome prediction systems becomes an urgent priority.

Two legal frameworks exist across the globe: the \textit{civil law system}, which assesses each case based on comprehensive codes and statutes, and the \textit{case law system}, where the interpretation and application of law heavily depend on precedent court decisions.
Most existing works were proposed for the civil law framework, including charge prediction, violated articles prediction, prison terms prediction, court decision prediction, and court view generation~\cite{paul-etal-2020-automatic, hu-etal-2018-shot, chen-etal-2019-charge, chalkidis-etal-2019-neural, DBLP:journals/corr/abs-2112-03414, ye-etal-2018-interpretable}. However, predicting case outcomes in case law systems presents unique challenges distinct from those in civil law: (1) it requires the identification of similar historical cases, and (2) meanwhile accounting for the evolution of legal principles over time.


\begin{itemize}[leftmargin=*]
    \item \textbf{Precedent Cases} In the case law system, the application of precedents plays a crucial role. To achieve accurate prediction of case outcomes, it is vital to identify past cases that exhibit similar legal principles, factual contexts, and key arguments. Moreover, how to effectively utilize the retrieved cases in the prediction of new case outcomes still requires further exploration.
    
    \item \textbf{Temporal Shift} One aspect that has not received sufficient attention in previous research is the temporal evolution of legal principles. We argue that it is crucial to not only comprehend the historical context and development of legal precedents but also to effectively capture and represent the temporal shifts of laws in predictive modeling.
\end{itemize}



To fill the gap, we proposed a new model named \method (\textbf{P}redict\textbf{I}ng \textbf{L}egal case \textbf{O}u\textbf{T}come) for case outcome prediction, which consists of two functional modules:

\begin{itemize}[leftmargin=*]
    \item \textbf{Case Retrieval} We initially train the module in an unsupervised manner to obtain text embeddings for cases. These embeddings are then used to query and select the most relevant precedent cases, which serve as additional inputs to our main model. 
    \item \textbf{Temporal Pattern Mining}  A temporal decay term is introduced to ensure the model captures the more recent patterns and explicitly learns to adapt to the temporal pattern change.
\end{itemize}


To facilitate this line of research, we established a new dataset named \textit{ECHR2023}, which was extracted from the European Court of Human Rights (ECHR) database\footnote{https://hudoc.echr.coe.int} with focusing on precedent cases and temporal concept shift. We evaluated the proposed \method model against state-of-the-art models on \textit{ECHR2023}. The experiment results show that \method substantially outperforms existing works in several metrics. The two modules in \method effectively improve the performance in different aspects.

In summary, the main contributions of this paper are as follows
\begin{itemize}[leftmargin=*, itemsep=0pt, labelsep=5pt]
    \item We highlight the issue of Temporal Pattern Shift in legal AI tasks. This problem is important but is usually ignored in most previous works.
    \item We propose a new method, \method, which can effectively handle Temporal Pattern Shift based on characteristics of the case law system. 
    \item We contribute a new dataset, \textit{ECHR2023}, for legal case outcome prediction. \method achieves state-of-the-art performance on \textit{ECHR2023}.
\end{itemize}

\section{Related Work}

\noindent\textbf{Legal Case Outcome Prediction}
on civil law framework has been well studied, mainly focusing on predicting whether the case description violates 
 existing legislation. Machine learning technologies, including multi-task learning \cite{feng_legal_2022}, few-show learning \cite{hu-etal-2018-shot, 10.1007/978-3-030-30490-4_19} has been adopted. Model explanation has been another focused \cite{jiang-etal-2018-interpretable, Zhong_Wang_Tu_Zhang_Liu_Sun_2020, 9627791, chen-etal-2019-charge, ye-etal-2018-interpretable, wu-etal-2020-de}.

In contrast, for  case law systems that heavily relies on judicial decisions of relevant precedent cases rather than solely on constitutional law when rendering final case outcomes,
there are relatively few studies  due to the lack of scarcity of large-scale, high-quality, and structured labeled data.  For instance, \cite{chalkidis-etal-2019-neural} utilize HIER-BERT to  first encode individual facts and then employ two layers of transformers to encode all the facts within a given case. \cite{chalkidis-etal-2021-paragraph} generate rationales through a text encoder sub-network that reads the text, a rationale extraction sub-network that identifies the most important words via a binary mask, and a prediction sub-network that classifies a hard-masked version of the text. They also incorporate rationale constraints as regularizers. \cite{paul-etal-2020-automatic} employ a fact encoding layer to encode facts and a charge encoding layer to encode charges. Subsequently, they use a Matching Layer, which incorporates an attention mechanism, to predict the final charges for each case.  \cite{malik-etal-2021-ildc} utilize a Hierarchical XLNet architecture  to predict case outcomes and  related interpretations. These efforts primarily focus on the classification of fundamental case outcomes. To the best of our knowledge, most of the existing works do not handle temporal pattern shift.\\

\noindent\textbf{Temporal Pattern Shift}
 arises due to changes in label distribution,  meaning, and etc. Existing research approaches this issue from different angles. For example, 
\cite{zhao-etal-2022-impact} analyzes the impact of temporal pattern shift on model explanations. \cite{sun2018concept} explored drift adaptation through transfer-based ensemble learning. Fan et al. \cite{fan2023dish} proposed to use two CONET networks to model the normalized parameters of historical and future windows separately, enabling normalization and prediction of future sequences. \cite{lu2023outofdistribution} introduced an out-of-domain representation learning approach utilizing adversarial learning to capture domain-specific segments and a domain-independent commonality representation. \cite{rosin_temporal_2022} introduced Temporal Attention and trained a transformer-based model with additional time-based inputs. In the legal field, \cite{chalkidis-sogaard-2022-improved} tackled temporal pattern shift in legal text classification by proposing Label-Wise Distributional Robust Optimization. This algorithm addresses temporal pattern shift stemming from class imbalance problems and enhances model robustness. However, the existing works are too general and are designed for simple scenes, so they do not perform well in adapting the more complex shift. There is still a lack of a comprehensive solution for legal models to adapt the shift in the legal field directly and naturally.

\section{ECHR2023 Dataset}
We build a novel dataset called \textit{ECHR2023} that takes the special challenges in legal case outcome prediction with case law. This dataset is derived from the most recent ECHR database. The primary focus of \textit{ECHR2023} is to investigate the issue of temporal pattern shifts in the legal domain.

\noindent \textbf{Data Acquisition and Processing}
The data extracted from the ECHR database is of low quality and contains a substantial amount of noise. The case documents are often excessively long, surpassing 2,000 words, and may consist of text in multiple European languages. As a result, the readability and quality of the text data are poor, posing difficulties for humans in comprehending the content of the cases.

Specifically, we prompt the large language model, \textit{gpt-3.5-turbo}\footnote{https://platform.openai.com/docs/models/gpt-3-5}, to process the raw data. The prompts guide the model to focus on the primary arguments in the case and summarize them into more concise points. Therefore, the output of LLMs will not introduce new information or fabrication to a case but rather retain the important parts of the original information. We employ LLMs to summarize original legal documents with the aim of simplifying the input and concentrating on identifying temporal pattern shifts. The prompts, example input-output of the LLM, and more details in the raw data processing can be found in Appendix~\ref{sec:llm_process}. The resulting sample is described by the following attributes: case ID, title of the case, outcome decision date of the case, corresponding legal article, and text description of the case. Following the processing results, we conduct a manual review of the generated summaries to ensure their quality and to eliminate any data that is obviously incorrect.




\noindent \textbf{Data Analysis}
Most of existing datasets are random split and ignore the temporal pattern change in the real world. We analyze the temporal pattern change in this dataset as follows: we perform outcome prediction using BERT \cite{devlin-etal-2019-bert} using both the random split and chronological split.  For random split (that does not consider case time), the performance of model training in Micro-F1 is $0.798$, and the testing performance is  $0.796$. While for chronological split data split that we train the model using previous cases and test on cases that happen later, the performance of model training in Micro-F1 is $0.737$ and the testing performance is  $0.677$, which shows the patterns learned from previous data cannot fully capture the signal in new cases.

\section{Methodology}
\subsection{\method Framework}

\begin{problem}[Legal Case Outcome Prediction with Case Law]
Given a set of $n$ chronological ordered cases $\mathbf{C} = \{C_i\}_i^n$, where each $C_i$ is represented by the text description of the case, the legal case outcome prediction aims at predicting whether a new case violates any legal article in $\mathbf{V} = \{V_1, V_2, ... V_n\}$. Here $V_i\in \{0, 1\}^L$ is the corresponding multi-hot label vector of the case $C_i$ violated articles, and $L$ is the total number of law articles. This task is a multi-label classification to decide the case $C_i$ violated law articles $V_j$.
\end{problem}


\begin{figure*}[t]
\centering
\includegraphics[width=0.8\linewidth]{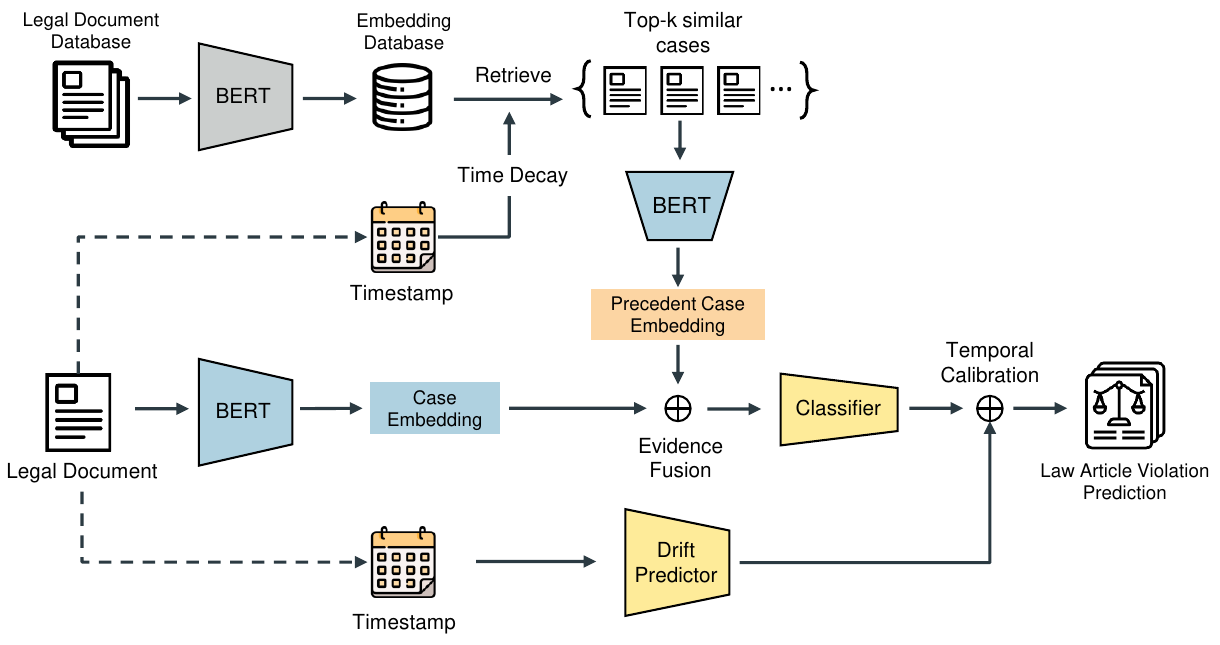}
\caption{The framework of our proposed model \method. PILOT has three modules: Relevant Case Retrieval, Case Encoder with Evidence Fusion, and Temporal Shift Mining. The Relevant Case Retrieval module retrieves relevant cases to use as references for outcome prediction. The Case Encoder with Evidence Fusion module encodes current cases with fact descriptions and relevant cases. The Temporal Shift Mining module adapts directly to temporal drift.}
\label{fig:framework}
\end{figure*}

We propose \method that primarily focuses on two distinct challenges in predicting legal case outcomes with case law: effectively identifying similar precedent cases and handling temporal pattern shift of legal principles. As illustrated by Figure~\ref{fig:framework}, \method consists of three modules: the Relevant Case Retrieval module that retrieves relevant cases as references for outcome prediction, the Case Encoder with Evidence Fusion module that uses encodes current case with fact description and relevant cases, and the Temporal Shift Mining module that is directly adapting to the temporal drift. We will now provide a detailed introduction to each of these modules.


\subsection{Precedent Case Retrieval}
In the case law system, precedent cases serve as crucial references that judges rely on when making decisions for new cases. In order to emulate this decision-making process, we develop a precedent case retrieval module that enhances case outcome prediction in two key aspects: (1) by providing augmented evidence for prediction and (2) by offering interpretability through the provision of evidence.

\noindent \textbf{Case Encoding} We execute contrastive learning based on a pre-trained language model on case documents only from training split of dataset. We suppose that we only have case documents in training split in the database at the beginning. The yielded model is then utilized for encoding all legal case documents in the database and the current case, preparing for similarity search. Formally, the contrastive learning is performed based on InfoNCE loss~\cite{gao-etal-2021-simcse}. Without the need for annotated labels, each case is passed into the BERT model twice within a batch, resulting in two different document embeddings $H_i^0$ and $H_i^1$ due to the randomness of the dropout layers~\cite{srivastava_dropout_nodate}. 

Within a batch, each pair $(H_i^0, H_i^1)$ is positive, and all the other pairs that  $(H_i^0, H_j^1)$ where $i \neq j$ are negative. The contrastive training objective $\ell_i$ is hence defined by:
\begin{equation}
\label{eq:simcse_objective}
\begin{aligned}
\ell_i = - \log \frac{e^{\mathrm{cos}(H_i^0, H_i^1)/ \tau }}{\sum_{j=1}^N e^{\mathrm{cos}(H_i^0,H_j^1)/\tau}},
\end{aligned}
\end{equation}
where $H_i$ is the document embedding of case $C_i$,  $H_i^0$ is the positive sample for $H_i$, $\tau$ is a temperature hyperparameter, and $\mathrm{cos}(H_i, H_j)$ measures the cosine similarity of the input embeddings.



\noindent \textbf{Case Retrieval} After training with Eq.~\eqref{eq:simcse_objective}, we encode all legal cases in the database into semantically meaningful document embeddings  $\mathbf{H} = \{H_1, H_2, ..., H_n\}$, which can be used to compute cosine similarities for case retrieval.

In this work, we put temporal constraints for the retrieval process. First, the retrieval is performed considering the timestamps of the target cases because a case cannot refer to any future cases. For a case $C_i \in \mathbf{C}$, we assign the similarity $\mathrm{sim}(C_i, C_j)=-1$ if $i < j$ to filter out future cases from the candidate pool.

Secondly, we also take into consideration the influence of temporal pattern shifts of legal principles, as recent cases often carry higher reference value in legal decision-making. Based on this insight, we design a variant of cosine similarity equipped with a temporal decayed function as
\begin{equation}\label{eq:similarity}
\mathrm{sim}(C_i, C_j) = \frac{\mathrm{cos}(C_i, C_j)}{1 + \frac{T_i - T_j}{\alpha \times |\mathbf{C}_{\mathrm{val}}|}},
\end{equation}
where $C_j$ is a candidate similar case, $\alpha$ is temporal decayed coefficient, and $|\mathbf{C}_{\mathrm{val}}|$ is the size of validation split in the dataset. We set the decaying unit as the size of the validation split because it is a time span from labeled data to the newest unlabeled data, which is also the length of validation data. When $\alpha = (T_i - T_j) / |\mathbf{C}_{\mathrm{val}}|$, the similarity score of $(C_i, C_j)$ will be half. As $\alpha$ decreases, the reference value of precedent cases will decrease faster. 

\subsection{Case Encoding with Evidence Funsion}

\noindent \textbf{Target Case Encoding}
To prepare legal case data for outcome prediction, the first step is to embed the case documents into contextualized representation.  To achieve this, we preprocess the legal document text data as follows: we convert the fact list to a piece of text by replacing all carriage return characters in the text with spaces, then use \textit{BertTokenizer} to conduct tokenization.

Next, the preprocessed legal document text data is passed into a pre-trained language model (PLM) for further processing. Here we choose \textit{legal-bert-base-uncased} \cite{chalkidis-etal-2020-legal}, which is pre-trained on different kinds of legal documents, enabling it to capture and understand the context and meaning of the text. For every case $C_i$, we pass it into the PLM and get the contextualized representation of the fact description $H_i \in \mathbb{R}^{d_t}$, where $d_t$ is the dimension of the last hidden layer in PLM.  We indicate this contextualized representation $H_i$ as the current case embedding $E_i$:
\begin{equation}
\begin{aligned}
E_i = H_i = PLM(C_i),
\end{aligned}
\end{equation}


The PLM takes the preprocessed legal document text data as input and generates a contextualized representation of the legal case text,  encapsulating the semantic and syntactic information of the legal fact description. It captures the relationships between words, phrases, and sentences, providing a rich representation of the text's meaning within the legal context.

\noindent \textbf{Evidence Fusion}
We use the target case $C_i$ to query all cases $\mathbf{C}$ to retrieve the top $k$ similar precedent cases according to similarity scores computed by Eq.~\eqref{eq:similarity}. We draw the evidence $\mathbf{R}_i = \{R_1, R_2, \dots, R_k\}$ from the retrieved cases, where $R_j = \{\mathrm{sim}(C_i,C_j), V_j\}$ includes the case result $V_j\in \{0, 1\}^L$ and the similarity score $\mathrm{sim}(C_i,C_j)$ of this relevant case. 

Based on the evidence $\mathbf{R}_i$ retrieved from precedent cases, we build the evidence embedding $E_i^r$ by:
\begin{equation}
\begin{aligned}
E_i^r = \sum_{j=1}^{k}\frac{e^{\mathrm{sim}(C_i,C_j)} \times V_j}{\sum_{j=1}^k e^{\mathrm{sim}(C_i,C_j)}}.
\end{aligned}
\end{equation}
where $E_i^r \in \mathbb{R}^{L}$. We concatenate current case embedding $E_i^c$ with relevant case embedding $E_i^r$ to get the input of the linear classifier layer for $C_i$ by:
\begin{equation}
\begin{aligned}
E_i = [E_i^c, E_i^r],
\end{aligned}
\end{equation}
where $E_i \in \mathbb{R}^{d_t + L}$. 

This approach allows the model to learn the relationship between relevant cases, leading to a better understanding of the factors influencing case outcomes. Moreover, it helps alleviate the impact of temporal pattern shift by providing a local perspective that captures the evolving nature of legal precedents.


\subsection{Outcome Prediction with Temporal Pattern Mining}
To further mitigate the temporal pattern drift when the model makes outcome predictions, we introduce a drift prediction module that mines the effect of timestamps to the final outcomes:
\begin{equation}
\begin{aligned}
\mathrm{Drift}_i = \mathrm{MLP}(T_i),
\end{aligned}
\end{equation}
where $\mathrm{Drift}_i \in \mathbb{R}^{d}$. $\mathrm{MLP}$ is a two-layer multi-layer perceptron, and the dimension of the hidden layer is $d$. We add the output $\mathrm{Drift}_i$ to the original prediction to get the final prediction:
\begin{equation}
\begin{aligned}
y^{\mathrm{final}}_i = y^{\mathrm{orig}}_i + \mathrm{Drift}_i,
\end{aligned}
\end{equation}
where $y^{\mathrm{orig}}_i$ is original output generated by the classifier and $y^{\mathrm{final}}_i \in \mathbb{R}^{L}$. The drift prediction module explicitly incorporates a global view by adapting to the temporal concept and learning from the entire timeline. By considering the evolution of legal precedents over time, this module effectively captures and adapts to the changes in the legal landscape, ensuring that the model remains robust and accurate in predicting case outcomes.

\subsection{Training and Loss Function}
In addition to the binary cross-entropy loss $\mathcal{L}_{BCE}$ used for the multi-label classification task, we add the drift loss $\mathcal{L}_{\mathrm{Drift}}$ to the model loss function. $\mathcal{L}_{\mathrm{Drift}}$ uses mean squared error loss to calculate the drift distance between original predictions and final predictions. The loss function of this model is defined as:
\begin{equation}
\mathcal{L} = (1 - \lambda) \mathcal{L}_{BCE} + \lambda \sum_{i=1}^{L}(y^{\mathrm{final}}_i-y^{\mathrm{orig}}_i)^2,
\end{equation}
where $\lambda$ is the weight that balances the two losses.

\section{Experiments}
In this section, we conducted extensive experiments to show the performance
of \method associated with more in-depth analysis. Universally, we report the average results of all models obtained by five runs with different random seeds, to ensure fair comparison. We use four metrics to evaluate the legal case outcomes: micro-F1, micro-Jaccard, micro-PR-AUC, and micro-ROC-AUC. More training details can be found in Appendix~\ref{sec:train}.

As for the availability of cases during training and evaluation, we strictly ensure that we do not use any later cases as references for the current case. During the training phase, all prior cases from the training set are available as precedents. At test time, all prior cases from both the training and test sets are available. In the contrastive learning of the case encoding model, we only use data from the training split of the dataset.


\begin{table*}[t!]
\centering
\aboverulesep=0ex 
   \belowrulesep=0ex 
\begin{tabular}{l|cccc}
\toprule
Method                     & F1                                 & Jaccard                           & PR-AUC                             & ROC-AUC                            \\ \hline
BERT                       & 0.675$\pm$0.005          & 0.509$\pm$0.005         & 0.498$\pm$0.004          & 0.795$\pm$0.011                 \\
HIER-BERT                  & 0.680$\pm$0.008          & 0.516$\pm$0.009         & 0.502$\pm$0.011          & 0.803$\pm$0.004          \\
BERT-LWAN                  & 0.655$\pm$0.012          & 0.488$\pm$0.014         & 0.477$\pm$0.009          & 0.782$\pm$0.017          \\
EPM-base                   & 0.657$\pm$0.012          & 0.490$\pm$0.013         & 0.482$\pm$0.014          & 0.781$\pm$0.006          \\
BERT+CL+kNN                & 0.679$\pm$0.006          & 0.514$\pm$0.007         & 0.502$\pm$0.006          & 0.793$\pm$0.015          \\
BERT+TemporalAttention     & 0.648$\pm$0.009          & 0.480$\pm$0.010         & 0.459$\pm$0.012          & 0.791$\pm$0.008          \\
LWDROV2                    & 0.694$\pm$0.013          & 0.531$\pm$0.015         & 0.511$\pm$0.016          & 0.830$\pm$0.011          \\
ChatGPT 5-shots    & 0.442                               & 0.284                              & 0.267                               & 0.818                               \\ 
\textbf{\method} (Ours) & \textbf{0.715$\pm$0.008} & \textbf{0.557$\pm$0.010} & \textbf{0.543$\pm$0.014} & \textbf{0.831$\pm$0.007} \\ \bottomrule
\end{tabular}
\caption{Experimental results. The best results are in bold. \method significantly outperforms all other methods in all metrics. $\pm$ represents standard deviation from five results of five different seeds.}
\label{tab:exp-res}
\end{table*}

\subsection{Baselines}
We consider the following baselines in evaluation.
\begin{itemize}[leftmargin=*, itemsep=0pt, labelsep=5pt]
    \item \textbf{BERT} \cite{devlin-etal-2019-bert} is a transformer-based \cite{vaswani2017attention} language model pretrained on large-scale web texts. We fine-tune and predict with the \textit{[CLS]} token of BERT.
    \item \textbf{HERT-BERT} \cite{chalkidis_neural_2019} is a hierarchical version of BERT. This model was proposed to predict legal judgment for long documents by first splitting and encoding raw law documents into multiple sentence embeddings, then fusing them with a two-layer Transformer model \cite{vaswani2017attention} to yield the document embeddings.
    \item \textbf{BERT-LWAN} \cite{chalkidis_empirical_2020} is Label-Wise Attention Network after BERT that was shown to be robust in multi-label classification. LWAN employs $L$ attention for $L$ labels to learn the semantics of label interpretation.
    \item \textbf{EPM-base} \cite{feng_legal_2022} is the variant of the state-of-the-art method on the CAIL2018 dataset. The original model, named Event-based Prediction Model (EPM) targets Chinese legal case outcome prediction, augmented by extra annotations about the legal event information. We remove the event extraction module in our experiments for fair comparison and refer the method to the name EPM-base.    
    \item \textbf{BERT+CL+kNN} \cite{su_contrastive_2022} is an advanced method for general purpose multi-label prediction. It is equipped with a k-nearest-neighbor model along with a multi-label contrastive learning objective for better multi-label classification performance.
    \item \textbf{BERT+TemporalAttention} \cite{rosin_temporal_2022} adds a time-aware self-attention module to the transformer model, which demonstrates superior performance in capturing temporal patterns when making predictions. In detail, it adds a time matrix to the attention weight to learn the impact of the temporal shift.
    \item \textbf{LWDROV2} \cite{chalkidis-sogaard-2022-improved} was proposed for legal text classification tasks. It employs Label-Wise Distributional Robust Optimization to mitigate class imbalance and temporal pattern shift problems. 
    \item \textbf{ChatGPT 5-shots} \cite{ouyang2022training} is based on the in-context learning capability of the \texttt{GPT-3.5-turbo} model. To be specific, we put the exemplar cases and their outcomes retrieved using our Precedent Case Retrieval module into the context, then prompt the language model to generate the outcome predictions.
\end{itemize}

\noindent\textbf{Evaluation Strategy.} To ensure that future information is not used in legal case outcome predictions, we partitioned the data chronologically. As a result, the training, validation, and test data consist of 8,138, 3,000, and 3,000 instances, respectively, ensuring a preserved time span between the sets. In addition, this chronological split enables the evaluation of models' adaptability to concept drift and reinforces temporal coherence. The dataset provides a substantial amount of validation and test data, contributing to its superior evaluation capabilities for legal case outcome prediction compared to existing alternatives. The statistics of the case outcomes are summarized in Table \ref{tab:echr-dist}.

\begin{table}[h!]
\aboverulesep=0ex 
   \belowrulesep=0ex 
\resizebox{.48\textwidth}{!}{%
\begin{tabular}{@{}l|ccc@{}}
\toprule
ECHR Articles                                    & Train & Dev. & Test  \\ \hline
Right to life                                & 432   & 180  & 188    \\
Prohibition of torture                       & 1,048  & 796  & 835   \\
Right to liberty and security                & 1,264  & 608  & 690   \\
Right to a fair trial                        & 4,969  & 1,165 & 1,081  \\
No punishment without law                    & 32    & 7    & 9      \\
Right for private and family life & 682   & 287  & 421   \\
 Freedom of religion  & 43    & 17   & 26     \\
Freedom of expression                       & 313   & 151  & 194   \\
Freedom of assembly         & 104   & 80   & 148    \\
Right to an effective remedy                & 1,202  & 506  & 520   \\
Prohibition of discrimination               & 170   & 48   & 61    \\
Derogation in time of emergency             & 4     & 9    & 10     \\
Individual applications                     & 58    & 46   & 60    \\
Examination of the case                     & 34    & 4    & 7       \\
Protection of property                    & 1,483  & 435  & 347   \\
Signature and ratification                & 5     & 11   & 21    \\ \bottomrule
\end{tabular}%
}
\caption{Label distribution of the ECHR2023 dataset.}
\label{tab:echr-dist}
\end{table}

\subsection{Result: Legal Case Outcome Prediction}
We report the main results of legal case outcome predictions in  Table~\ref{tab:exp-res}. From the table, we observe that our method outperforms other methods by a large margin in four metrics, especially over the methods that do not explicitly consider the temporal pattern shifts in legal case outcomes. 


In addition, our method improves the micro-F1 by 2.74\% than the previous state-of-the-art method of legal outcome prediction, LWDROV2. The reason is that LWDROV2 is a general label-wise robust method that does not solve temporal shifts directly. By contrast, our method employs a time-aware drift prediction module and augments the predictions with precedent cases.


It is noteworthy that ChatGPT 5-shots exhibits lower performance when compared to other prediction models based on supervised learning. In many instances, ChatGPT refuses to provide predictions, leading to limitations in its ability to make accurate determinations. Consequently, there remains the potential for further advancements in general-purpose generative large language models for predicting legal outcomes.


\begin{table}[t!]
\aboverulesep=0ex 
   \belowrulesep=0ex 
\centering
\begin{tabular}{@{}l|cc@{}}
\toprule
Method                     & \multicolumn{1}{c}{F1} & \multicolumn{1}{c}{$\bigtriangledown$} \\ \hline
\textbf{\method} & \textbf{0.712}          & \textbf{-}                                          \\ 
w/o relevant case retrieval             & 0.701                   & -0.011                                                \\
w/o temporal pattern handling        & 0.697                   & -0.015                                                \\
w/ law article semantics   & 0.705                   & -0.007                                                \\ \bottomrule
\end{tabular}
\caption{Results of ablation study. Relevant case retrieval and temporal pattern handling bring improvement to the model respectively, while incorporating law articles semantics has a performance drop. $\bigtriangledown$ means the performance drop comparing with the method \method.}
\label{tab:ablation}
\end{table}

\subsection{Result: Ablation Study}
We performed an ablation study to evaluate the impact of the relevant case retrieval module and the temporal pattern handling module on the overall performance of our model. Table~\ref{tab:ablation} presents the results of this study, highlighting how these two modules contribute to the improvement of the base model in distinct ways.

Additionally, we explored the incorporation of law article semantics into the model, using techniques such as law side attention or similar approaches employed in previous methods. Surprisingly, our findings indicated a decrease in performance when integrating law article information into our model. This observation is supported by the results in Table~\ref{tab:exp-res}, where both the EPM-base and BERT-LWAN models, which incorporate law article information, exhibited inferior performance compared to BERT alone. We think one reason incorporating law articles undermines the performance is that the content and interpretations of law articles change as time goes on. It will influence model prediction without considering the time factor.

\begin{table*}[t!]
\centering
\aboverulesep=0ex 
   \belowrulesep=0ex 
\resizebox{\linewidth}{!}{
\begin{tabular}{@{}l|clcc@{}}
\toprule
Case                      & case id    & main text (selected sentences)                                                                                       & violated articles & similarity  \\ \hline
\textbf{Current Case}      & \textbf{001-199268} & \textbf{the applicant complained about the lack of effective remedy in domestic law}                & \textbf{["13", "6"]}   & \textbf{-}                \\ \hline
Precedent Case 1 & 001-195868 & the applicant expressed concerns about the lack of effective remedies in domestic law & {[}"3", "13"{]}   & 0.597            \\
Precedent Case 2 & 001-189950 & applicant complained about inadequate detention conditions                                 & {[}"3", "13"{]}   & 0.560            \\
Precedent Case 2 & 001-198818 & applicant complained about the excessive length of civil proceedings                       & {[}"13", "6"{]}   & 0.421            \\
 Precedent Case 3 & 001-199269 & complaint concerns the length of administrative proceedings regarding social benefits  & {[}"6"{]}         & 0.380            \\
Precedent Case 4 & 001-198820 & the applicant complained about the excessive length of his pre-trial detention            & {[}"6", "5"{]}    & 0.364            \\ \bottomrule
\end{tabular}}
\caption{An example of similar case retrieval results. }
\label{tab:casestudy}
\end{table*}

\subsection{Result: Qualitative Case Study for Case Retrieval}
The relevant case retrieval module is utilized for retrieving the top $k$ precedent cases that are relevant to the target case. In Table~\ref{tab:casestudy}, we present an example of the retrieval results. It is evident from the table that these retrieved cases exhibit semantic relevance to the target case. Furthermore, the violated articles mentioned in the retrieved cases are closely related and encompass the violated articles of the target case, indicating a comprehensive coverage of relevant legal provisions. Therefore, it demonstrates the effect of the case retrieval process from a qualitative perspective.


\begin{figure}[h!]
\centerline{
\resizebox{8cm}{!}{
\includegraphics{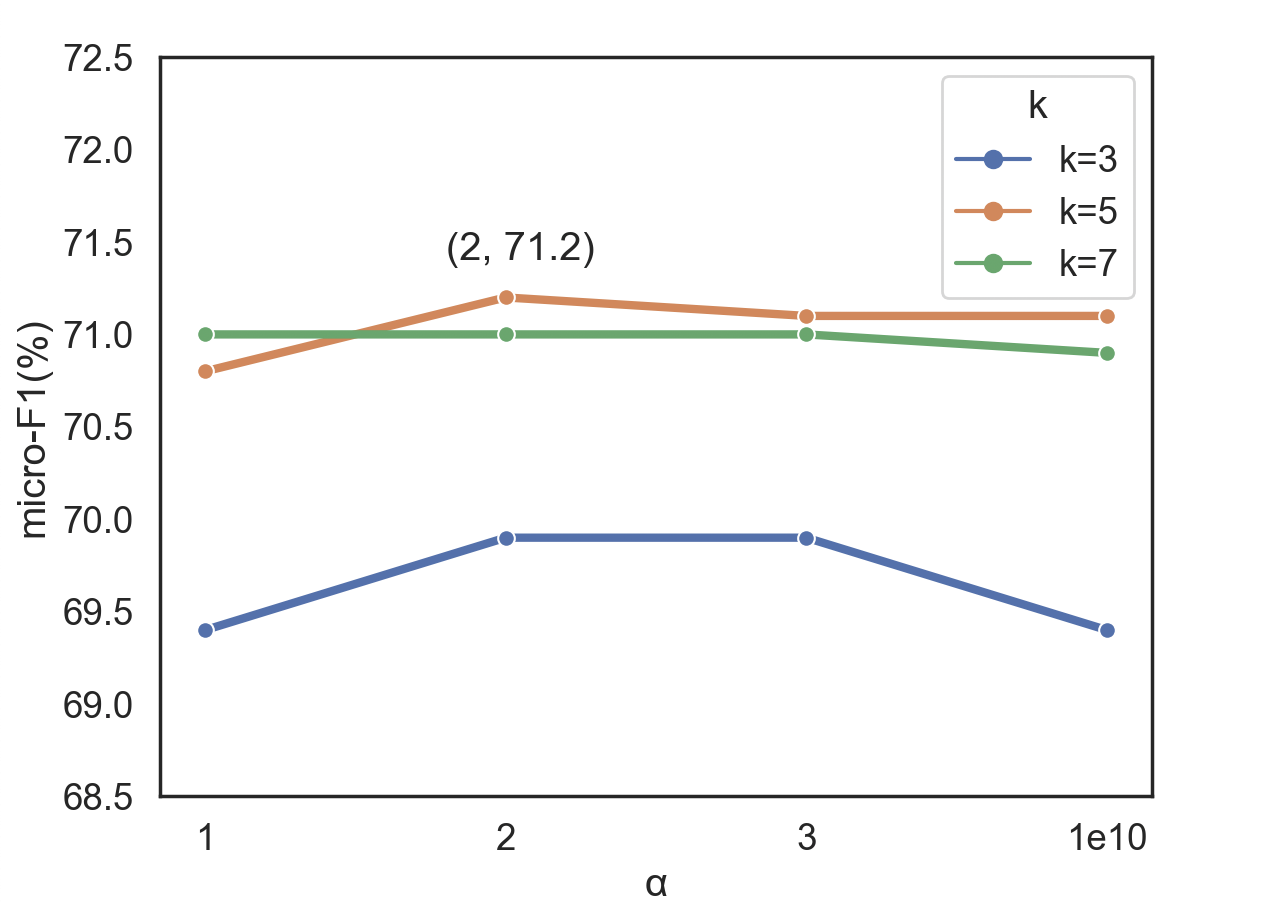}
}
}
\caption{Hyperparameter analysis of $k$ and $\alpha$ in the relevant case retrieval module. When $k$ equals 5 and $\alpha$ equals 2, the model achieves the best results. When the value of $\alpha$ is 1e10, it indicate an extreme condition that implies the absence of temporal decay in the computation of the similarity score}
\label{fig:kanda}
\end{figure}

\subsection{Result: Hyperparameter Analysis for Case Retrieval Module}
The relevant case retrieval module encompasses two hyperparameters. The first parameter, denoted as $k$, determines the number of top relevant precedent cases to be retrieved. The second parameter is the coefficient $\alpha$ associated with the temporal decayed function in Eq.~\eqref{eq:similarity}. The experimental results, presented in Table~\ref{fig:kanda}, shed light on the impact of these hyperparameters.

From the results, we conclude that including only three reference cases can introduce noise and lead to a decrease in performance, as it fails to retrieve the correct relevant cases effectively. However, utilizing five or seven reference cases demonstrates improved robustness compared to three cases. Notably, setting the value of $\alpha$ to 1e10 is an extreme condition that implies the absence of temporal decay in the computation of the similarity score. The results indicate that incorporating the time-decayed function brings about some improvement over the original approach. Empirically, we find setting $\alpha \in [1,10]$ yields the optimal results.


\begin{figure}[h!]
\centerline{
\resizebox{8cm}{!}{
\includegraphics{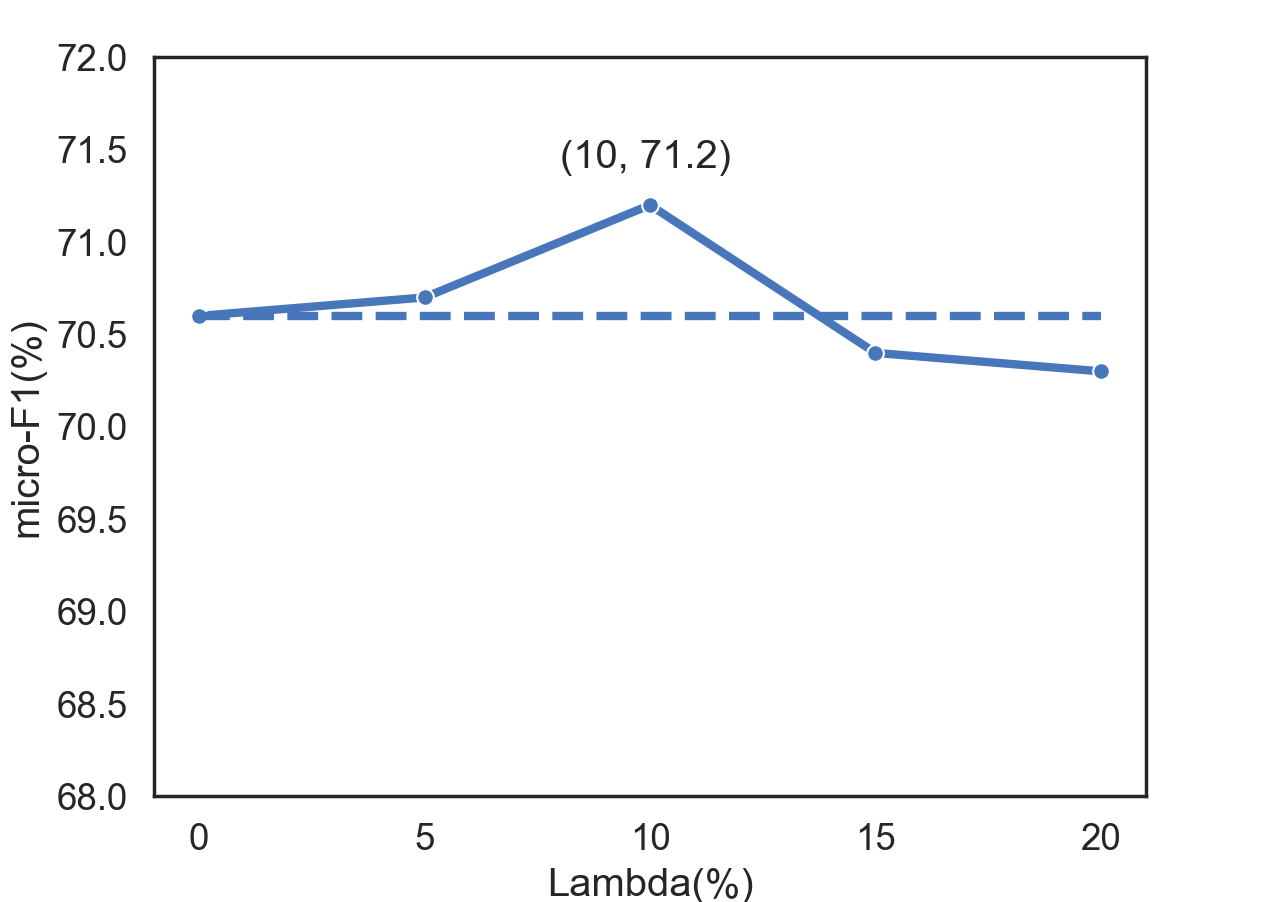}
}
}
\caption{Hyperparameter analysis of lambda which is the weight of drift loss. When $\lambda$ equals 0.10, the model achieves the best results.}
\label{fig:lambda}
\end{figure}

\subsection{Result: Hyperparameter Analysis for Training Objective}

To assess the impact of varying drift loss weights ($\lambda$), we conducted evaluations using different values. The results are presented in Figure~\ref{fig:lambda}. It is evident from the table that the inclusion of the drift loss contributes to improved model training and overall performance. Notably, the best value for $\lambda$, which balances the weighting between $\mathcal{L}_{\textrm{BCE}}$ and $\mathcal{L}_{\textrm{Drift}}$, is found to be 0.10. The $\lambda$ value of 0 indicates the exclusion of the drift loss from the model. Conversely, assigning a large value to $\lambda$ can have a detrimental effect on the model's performance.

\section{Conclusion}
In conclusion, this paper introduces the \method model to tackle the challenges associated with predicting case outcomes in case law systems. Through our experiments, we have demonstrated the superior accuracy of our model in predicting case outcomes compared to existing methods. This improvement can be attributed to the identification of similar cases and the effective handling of temporal pattern changes.

Moreover, our proposed model goes beyond enhancing the accuracy of legal case predictions. It also offers valuable insights into legal reasoning and the evolution of legal principles. Precedent cases hold significant importance within the case law legal framework. It is worth noting that many previous works have primarily focused on the civil law system, which differs from the case law system. By analyzing and leveraging precedent cases, our model provides a deeper understanding of the underlying legal principles and their application.



\section*{Limitations}
Deciding the outcome of legal cases is a very complex process in the real world. In this paper, we simplify many settings in real court scenarios to facilitate our research. The proposed model \method is a preliminary work in legal case outcome prediction, which might serve as a baseline for future investigation. The goal of designing the \method model is to highlight and alleviate the temporal pattern shift. There are many bias problems that need to be eliminated, and the model needs better interpretability to give reliable outcomes. It cannot be applied in the real world directly. Here are some ways to enhance the capability of \method before its application:

\begin{itemize}[leftmargin=*, itemsep=0pt, labelsep=5pt]
    \item More factors should be considered when designing a precedent case retrieval module. Currently, relevant cases are determined based on semantic similarity alone. However, relevant cases may not always be entirely semantically similar. Additionally, differences in factual details among cases can lead to different legal outcomes. Therefore, a more robust retrieval module with more retrieval factors should be developed if PILOT is to be applied in real-world scenarios.
    \item We need to further eliminate bias issues of \method before applied in real life.
    \item The model should prioritize better interpretability in order to provide reliable outcomes, given the need for transparency in the legal domain. For example, we can add a generation module let \method generate some explanation of its judgement.
    \item Legal outcomes should not be determined by a single model alone. Instead, a Mixture-Of-Experts approach can be employed, utilizing multiple instances of \method with varying hyperparameters, to perform ensemble learning and generate diverse results. After a voting process, the results can be more impartial.
    \item The model can benefit from incorporating more information from the case. Currently, only the factual section of the case is utilized, but additional information could be included to improve the model's performance.
\end{itemize}

\section*{Ethics Statement}
\noindent\textbf{Accuracy and Transparency.} We are committed to ensuring the accuracy of our predictions to the best of our abilities. We will maintain transparency about the methodologies, data sources, and algorithms used in our prediction models. We understand the profound implications of our work and strive to prevent any potential harm caused by inaccurate predictions. \\

\noindent\textbf{Fairness and Impartiality.} We pledge and strive to ensure our prediction models do not perpetuate or amplify any form of bias or discrimination. We will regularly audit our models to detect and mitigate any unfair bias, ensuring our predictions are objective and impartial. \\

\noindent\textbf{Respect for Privacy and Confidentiality.} We will strictly adhere to all applicable laws and regulations concerning data privacy and confidentiality. We will only use data that has been lawfully and ethically obtained, ensuring the privacy of all individuals involved is respected. \\

\noindent\textbf{Accountability.} We acknowledge our responsibility for the predictions made by our models. We will continually monitor and refine our models to ensure their reliability and validity. \\

\noindent\textbf{Legal Compliance.} We understand the significance of legal regulations and standards in our work. We will ensure full compliance with all relevant legal and professional guidelines in our legal outcome prediction task.



\bibliography{anthology,custom}

\appendix


\newpage 

\section{Training Details}
\label{sec:train}
In the model training, we fine-tune on \textit{legal-bert-base-uncased} \cite{chalkidis-etal-2020-legal}. AdamW optimizer \cite{DBLP:journals/corr/abs-1711-05101} was used to optimize the parameters of the model during the training. We apply differential learning rates. The learning rate of the final linear classifier is set to 1e-3, while others are all set to 1e-5. The Dropout \cite{srivastava_dropout_nodate} rate after the PLM output is set to 0.2. The batch size in each training step is
set to 8. In training, we set an early stop strategy with 2 epochs. We use micro-F1 as monitoring indicators in our early
stop strategy. We train \textit{CaseSifter} with all data 3 epochs.

The code implementation of our model is mainly written using
PyTorch \cite{paszke_pytorch_nodate} library, and the pre-trained model is loaded using Transformers \cite{wolf_transformers_2020} library. In addition, model training and evaluation were conducted on one NVIDIA GeForce RTX 3090.

\section{Raw Data Processing with LLMs}
\label{sec:llm_process}
We use large language models (LLMs) to process raw data. The original document is lengthy and redundant. Our summarization target is the FACT section of case documents. We employ multiple regular expressions to filter out only the FACT section from the case documents and then input them into the LLM. We ensure that the input data does not contain any other parts of the case documents, which may leak information about the results. We prompt the \textit{gpt-3-5-turbo} model to get output as processed data of a long document of one legal case. We utilized the default hyperparameters, setting the \textit{temperature} to 1 and the \textit{repetition\_penalty} to 0. The maximum sequence length of the output is set to 512 tokens to ensure compatibility with BERT.

\begin{figure*}[h!]
\centerline{
\resizebox{17cm}{!}{
\includegraphics{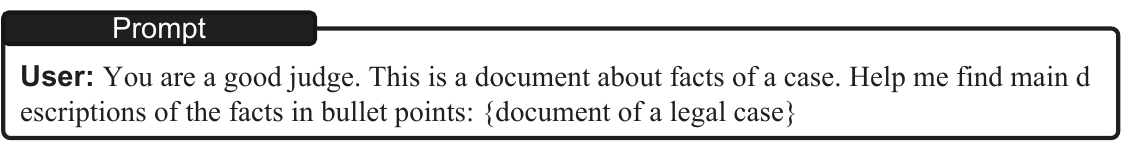}
}
}
\caption{The final selected prompt. We also prompt model by telling \textit{you are a good judge}.}
\label{fig:prompt}
\end{figure*}

We have tried several prompts and select prompt according to summary performance of the model. The final selected prompt is shown in Figure~\ref{fig:prompt}. In our prompts, we guide the model to focus on the primary arguments in the case and summarize them into more concise points. Therefore, the output of LLMs will not introduce new information or fabrication to a case but rather retain the important parts of the original information. In this case, we can minimize the problem of hallucinations caused by generative language models as much as possible. We also acknowledge that this method will cause potential semantic loss in new dataset, but it can increase the model inference speed and improve readability of original case documents. 

We manually check the data quality from the LLM output. We review about dozens of samples of data. We observe that it do not introduce any new fabricated facts in the output, and indeed summarizes some key points of the case, which meets our expectations. We have also conducted experiments to compare these aspects. Our results show that using the baseline results of ChatGPT processed content only leads to a 0.5\% decrease in performance than original lengthy documents, but significantly increases the training speed in later stages.

\begin{figure*}[h!]
\centerline{
\resizebox{17cm}{!}{
\includegraphics{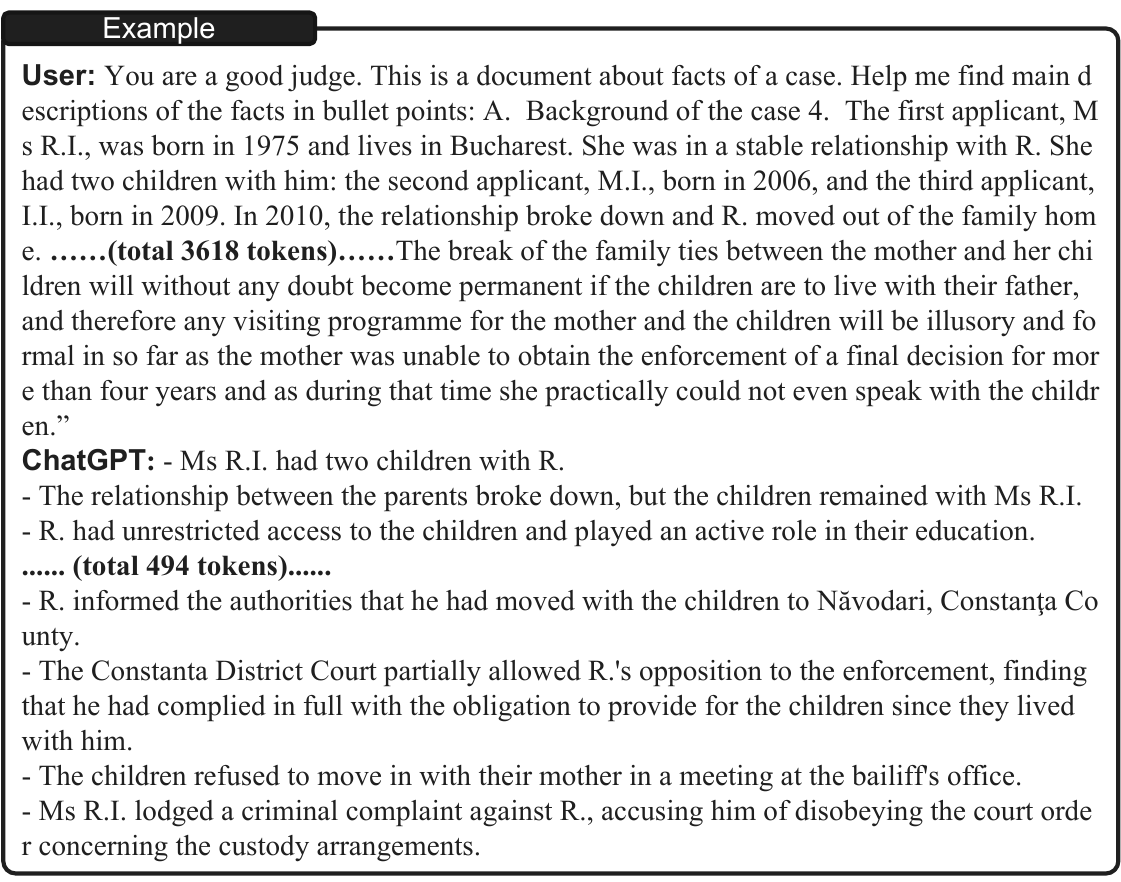}
}
}
\caption{An example input and output of the LLM about data \textit{001-187931}. The original document has 3618 tokens totally. It reduces to 494 tokens after extracting important points of a legal case.}
\label{fig:ioexample}
\end{figure*}

An example input and output of the LLM in data processing is shown in Figure~\ref{fig:ioexample}.

\end{document}